\documentclass[10pt,journal,compsoc]{IEEEtran}

\ifCLASSOPTIONcompsoc
  \usepackage[nocompress]{cite}
\else
  \usepackage{cite}
\fi

\usepackage{hyperref}

\ifCLASSINFOpdf
  \usepackage[pdftex]{graphicx}
  \graphicspath{{../pdf/}{../jpeg/}}
  \DeclareGraphicsExtensions{.pdf,.jpeg,.png}
\else
  \usepackage[dvips]{graphicx}
  \graphicspath{{../eps/}}
  \DeclareGraphicsExtensions{.eps}
\fi

\usepackage{amsmath}
\usepackage{algorithmic}

\usepackage{array}

\ifCLASSOPTIONcompsoc
  \usepackage[caption=false,font=footnotesize,labelfont=sf,textfont=sf]{subfig}
\else
  \usepackage[caption=false,font=footnotesize]{subfig}
\fi

\usepackage{stfloats}

\ifCLASSOPTIONcaptionsoff
  \usepackage[nomarkers]{endfloat}
 \let\MYoriglatexcaption\caption
 \renewcommand{\caption}[2][\relax]{\MYoriglatexcaption[#2]{#2}}
\fi

\usepackage{url}

\def\BibTeX{{\rm B\kern-.05em{\sc i\kern-.025em b}\kern-.08em
    T\kern-.1667em\lower.7ex\hbox{E}\kern-.125emX}}

\usepackage{amsmath,amssymb,amsfonts}
\usepackage{xcolor}
\usepackage{enumerate}
\usepackage{enumitem}

\begin{document}
\title{Confidence estimation of classification based on the distribution of the neural network output layer}

\author{Abdel Aziz Taha *1,
        Leonhard Hennig *2,
         Petr Knoth *1

\IEEEcompsocitemizethanks{\IEEEcompsocthanksitem *1: Research Studios Forschungsgesellschaft, Data Science, RSA FG, Vienna, Austria, E-mail: abdel\_aziz\_taha@hotmail.com.
\IEEEcompsocthanksitem *2: Deutsches Forschungszentrum für Künstliche Intelligenz (DFKI). Alt-Moabit 91c, 10559 Berlin, Deutschland.

}}

%\markboth{IEEE Transactions on Knowledge and Data Engineering,~Vol.~xx, No.~xx, August~2019}%
%{Shell \MakeLowercase{\textit{et al.}}: Bare Demo of IEEEtran.cls for Computer Society Journals}

\IEEEtitleabstractindextext{%
\begin{abstract}
One of the most common problems preventing the application of prediction models in the real world is lack of generalization: The accuracy of models, measured in the benchmark does repeat itself on future data, e.g. in the settings of real business. There is relatively little methods exist that estimate the confidence of prediction models. In this paper, we propose novel methods that, given a neural network classification model, estimate uncertainty of particular predictions generated by this model. Furthermore, we propose a method that, given a model and a confidence level, calculates a threshold that separates prediction generated by this model into two subsets, one of them meets the given confidence level. In contrast to other methods, the proposed methods do not require any changes on existing neural networks, because they simply build on the output logit layer of a common neural network. In particular, the methods infer the confidence of a particular prediction based on the distribution of the logit values corresponding to this prediction. The proposed methods constitute a tool that is recommended for filtering predictions in the process of knowledge extraction, e.g. based on web scrapping, where predictions subsets are identified that maximize the precision on cost of the recall, which is less important due to the availability of data. The method has been tested on different tasks including  relation extraction, named entity recognition and image classification to show the significant increase of accuracy achieved. 
\end{abstract}

\begin{IEEEkeywords}
Confidence estimation, prediction certainty, model evaluation, generalization.
\end{IEEEkeywords}}

\maketitle

\IEEEdisplaynontitleabstractindextext

\IEEEpeerreviewmaketitle

\section{Introduction}

Neural Networks (NN) gained a great importance in classification tasks. However, like other algorithms, NN models suffer from the generalization problem, that is, when applied on future data, NN models do not always repeat their accuracies that have been measured on the test dataset. An additional problem of NN models is that they always attempt to provide classifications with high probabilities (softmax values), even when these models are not (well) trained or when the example is out-of-distribution, i.e. they are never aware when they fail \cite{Hendrycks2016}. These shortages make them difficult to be used for decision making in business einvironments or in critical task like in medical applications \cite{Amodei2016}. Being not aware of the reliability of classification models because of missing generalization is a problem that hinders using such models especially in critical applications where important decisions are to be made, like in business and medicine. In such situations, the knowledge about the reliability and confidence of outcomes of the models are of great importance.    

We present here some examples of notes reported in research on the problem of uncertainty in Knowledge Discovery to raise the importance of confidence estimation: "Traditional IE and KD techniques are facing new challenges of dealing with uncertainty of facts extracted from heterogeneous data sources from different documents, languages and modalities on the WWW" \cite{uncertainty_reduction_IE}. "Veracity stresses the importance of data quality and level of trust due to the concern that many data sources (e.g. social networking sites) inherently contain a certain degree of uncertainty and unreliability." \cite{big_data_in_SCM}. "While most of the reviewed studies focus on designing and evaluating a mathematical model that takes a number of uncertainties and risks into account, there is less focus on establishing and analysing the applicability of the proposed models" \cite{SCM_and_AI_2019}. "While uncertainty estimation in neural networks is an active field of research, the current methods are rarely adopted. It is desirable to develop a method that does not create additional computational overhead "\cite{Mozejko2018InhibitedSF}. These notes and many others emphasize the problem of uncertainty, which raises the awareness of the need of methods quantifying uncertainty and confidence in AI models.

In the typical case, a neural network has an output layer with a number of nodes equal to the number of classes. Sometimes with an additional node representing the negative class, i.e. a class representing objects not belonging to any of the predefined classes. Given an object as input, a NN model as a classifier responds with an outcome on the output layer in form of values, each of them corresponds to one of the nodes in the output layer. Each of this values, which are called logits, corresponds to a class and reflects the likelihood that the example object (input) is assigned to this class. In case of existence of a negative class, the value reflects the likelihood of the object being assigned to none of the given classes. The name \textit{logits} refers to the logit activation function that normally produces these values. However, for simplicity, for the remainder of this paper, we will denote these values as \textit{logits} and the corresponding layer as \textit{logit-layer} regardless of the activation function used to produce them. 

The most common way to classify objects based on the logit layer is to simply assign an object to the class with the largest logit value without taking the other logit values consideration. In this paper, we use the distribution of the logit values to infer information about the confidence of the model when performed to classify a particular object. 

In Particular, we provide methods that enable filtering predictions by selecting those that are most probably correctly classified. This is done based on two things: First, confidence estimation functions are developed that, based on the logit values, calculate a confidence measure for a particular prediction. Second, a method for calculating a threshold for a particular model, based on which the predictions are filtered according their confidence values. 

The methods provided in this paper are useful and especially designed as a tool to increase confidence in those cases where knowledge is extracted from the web. More specifically, it is designed to filter classifications performed on data that is scraped from the web to extract knowledge from, e.g. to build knowledge graphs. In those cases, data is normally unstructured, heterogeneous and noise. These conditions promote two problems: The low accuracy due unclean noisy data and less generalization due to heterogeneity. The methods provided in this paper help to increase precision by filtering out classifications that are probably miss-classified on cost of recall, which is in those cases less important since there are a lot of data in the web.      

There are two main contributions of this paper: First, we propose confidence estimation functions that provide an estimate of how confident the NN model is, when applied on of a particular object. This is done based on the distribution of the values of the logit-layer corresponding to the object being classified. Second, for a given model and a given confidence level, we provide a method for finding a threshold based on which the predictions can be divided into two subsets, one subset meets the given confidence level, which call \textit{exploit} subset and another one of low quality and confidence, which we call \textit{waste} subset.   

The remainder of this paper is organized as follows: In Section~\ref{sec:related_work}, we present related work in confidence and reliability of NN models. After introducing some notations in Section\ref{sec:notation}, we present two novel confidence estimation functions in Section~\ref{sec:confidence_functions} and a method for defining a threshold for filtering classifications with regard to a given confidence level in Section~\ref{sec:confidence_thresholds}. Finally, the proposed methods are discussed and evaluated in Section~\ref{sec:evaluation}.

\section{Related work}
\label{sec:related_work}

The research most related to our work might be Hendrycks and Gimpel (\cite{Hendrycks2016}), where the authors show that the maximum of the softmax outcome in a neural network gives information about the uncertainty of the model. In particular, they show that correctly classified examples tend to have a relatively larger maximum in the softmax outcome. They assessed their finding using several tasks in computer vision, natural language processing, and automatic speech recognition, Thereby, they defined a simple baseline for utilizing the softmax outcome for estimating the uncertainty of a neural network classification models. However, to differentiate our work, first Gimpel et al. (\cite{Hendrycks2016}) apply their methods on top of the softmax outcome, although there are much research, such as \cite{9412540} \cite{SubramanyaSB17} \cite{abs-1811-08577}, stating that the softmax tends to provide very high changes in the probabilities for small changes in the logits due to the exponential functions in the definition of the softmax and this property makes it poor for predicting confidence \cite{Hendrycks2016}. In contrast, we build our methods rather on the original values, i.e. the logit values, which are the input of the softmax rather than the output and they are less sensitive in this regards. Second, while Gimpel et al. (\cite{Hendrycks2016}) only state the general relation between the maximum of the softmax and the uncertainty, we provide functions that quantify the uncertainty of a particular prediction based on the distribution of the corresponding logit values. Third, we provide a method to interpret this uncertainty in in the context of confidence levels, in particular it calculates a threshold to filter predictions such that they meet a given accuracy level in relation to a required confidence level.

Inspired by the research above, Tran et al. (\cite{NIPS2019_9607}) introduce a module that infers Neural Network Uncertainty, which can be used used as a building block that replaces the common  neural network layers. This module, which they call Bayesian layer, is capable to discover uncertainty based on the distributions of values over layers and activation functions using the concept of Bayesian Neural Network Layers. While this approach suggest an entire change the architecture of the neural network, our proposed methods do not require any change because they build on the common logit layer, i.e. they infer uncertainty based on the distribution of the logit values in the existing neural networks as they are.

Mozejko et al. (\cite{Mozejko2018InhibitedSF}) provide a method to quantify the uncertainty of a NN classifier by replacing the standard softmax function in the output layer with a new modified softmax function that they called the inhibited softmax function. In particular, the softmax function is modified such that it infers uncertainty of the model by capturing the cross-entropy loss function values in the training phase. The 

Other research, such as Blundell et al. \cite{Blundell2015WeightUI},  Louizos and Welling\cite{pmlr-v70-louizos17a}, Malinin and Gales \cite{Malinin2018PredictiveUE}, Wang et al. \cite{pmlr-v80-wang18i}, Hafner et al. \cite{Hafner2018ReliableUE}, try to tackle the problem of uncertainty by inferring the model confidence from the distribution over the models’ weights, an approach that has been inspired by the Bayesian approaches suggested by Buntine and Weigend \cite{Buntine1991BayesianB}. However, as there approaches aim to infer the model uncertainty, they are not designed to estimate and quantify uncertainty at level of a particular prediction, which our methods do.

\section{Methods}
\label{sec:methods}
In this section, we propose methods for estimating the confidence of classification done by a neural network (NN) based on the distribution of the output layer. 

\subsection{Notation}
\label{sec:notation}
Given a neural network trained to produce the classification model $\mathcal{M}$, let  $x = \{x_1, ....., x_n\}$ be a set of objects classified by the model $\mathcal{M}$ into $r$ predefined classes $c =\{ c^1,..,c^{r} \}$. Furthermore, let $L_i=\{L_i^1,...,L_i^r\}$ be the logit values of the output layer corresponding to the objects $o_i$. In this paper, the term \textit{classification} denotes in relation to a single object the process of assigning the object to a particular single class and in relation to a dataset assigning each of the objects in the dataset to a particular single class. Given the object $o_i$ classified by the model $\mathcal{M}$, we denote the largest corresponding logit $\hat{L_i}$ and the related class $\hat{c}$. Analogously, we denote the smallest corresponding logit $\check{l^i}$ and the related class $\check{c}$ 

In a common NN, the logit value $l_i^j$ represents the likelihood that the object $o_i$ belongs to the class $c^j$. Note that we intentionally say the likelihood because the logit values are not necessarily probabilities and they don't necessarily sum to one. The range and distribution of the logits depend on different factors like the activation function used for the output layer as well as on the technology and the implementation of the NN. However, the logits always tend to reflect the relative likelihoods of the object being in the corresponding classes. Also, note that the most common way to apply a binary classification is to select the class with the largest logit, i.e. $\hat{c}$.

\subsection{Confidence Functions}
\label{sec:confidence_functions}
A confidence function $conf(x)$ in the sense of this paper is a function that provides a value reflecting the confidence of the prediction $x$, i.e. a measure that reflects the probability $x$ being a true prediction.

In this section, we propose two functions of the logit values related to a particular classification to be estimators of the confidence of the corresponding classification, two implementations of the function $conf()$.

\begin{figure}[t]
\centering
    \includegraphics[width=\linewidth]{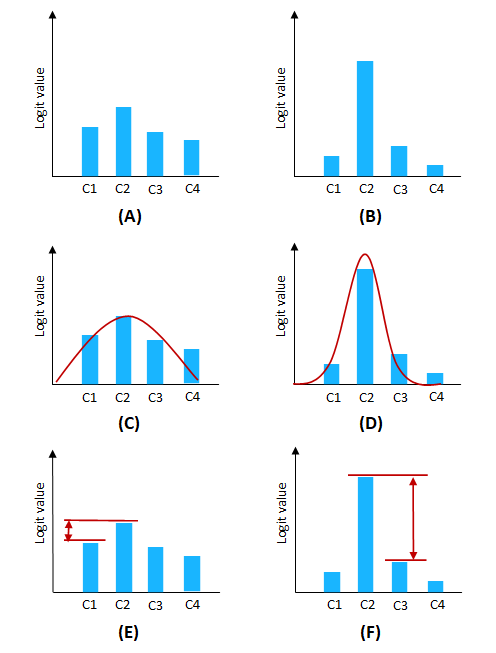}
    \caption{(A) and (B) are exemplary logits of two different classifications, where in both cases each of the objects encircled is assigned to class C2. In (A) the logit of the winner class has weak difference to the other classes, which would mean that in (A) the winner class is not entirely sure. In contrast, the winner in (B) is significantly distinct from the other classes, which would mean that the winner class is more sure. (C) and (D) apply the kurtosis function as a measure of recognizing the winner class. (E) and (F) applies only the difference between the two top winners and ignore the rest of the logits.}
    \label{fig:function_illustration}
\end{figure}

Before we present the functions, let's motivate the idea behind them. To illustrate how the distribution of the logits reflects the confidence, let's consider two exemplary logits distributions corresponding to two different classifications with four classes (C1, C2, C3 and C4) as shown in Figure~\ref{fig:function_illustration} (A) and (B). Both classifications suggest the class C2 as label since C2 has the maximum logit value in both cases. However, in (A), the differences between the logit value of the winner class C2 and those of the other classes are not significant. In contrast, C2 in (B) is a winner with distinction since there is are larger differences between the winner. The motivation here is that the more difference between the two largest logits, the more distinguished they are which is in turn an indicator that the model is more confident.   

The question now is how to measure the distinction of the winner class based on the logits? Figure~\ref{fig:function_illustration} (C) and (D) applies the function kurtosis as a measure of distinction, because higher kurtosis means a large value (peak) representing the winner class compared with many smaller values (tails) representing the other classes. Figure~\ref{fig:function_illustration} (E) and (F) applies the difference between the two highest classes (the winner and next best winner) as a measure, whereby rest of the logit values are ignored.

Based on this motivation, we propose the two following functions as confidence measures. Both functions are applied on the logits $L$ corresponding to an object $x$ which has been assigned to class c to estimate the confidence of the classification $c$.  
The first function $KRT$ is the kurtosis of the logits, which is by definition the fourth moment of the distribution. $KRT$ is defined as

\begin{equation}
   conf(x) = KRT(L_x) = E\Big[\Big( \frac{L_x - \mu}{\sigma}\Big)^4\Big]
   \label{eq:krt}
\end{equation}

The second function $WDF$ is the normalized difference between the logits of the winner and second winner classes. $WDF$ is defined as
\begin{equation}
   conf(x) = WDF(L_x) = \frac{Max1(L_x) - Max2(L_x)}{abs(Max1(L_x) + Max2(L_x))}
   \label{eq:wdf}
\end{equation}

where Max1 is the largest and Max2 is the second largest logit value. Note that $WDF$ is always in between the range $[0,1]$ regardless of the range of the logits. When the two largest logits are equal ($Max1= Max2$), $WDF$ provides a zero confidence. When $Max2$ is zero, this means that all other logits are zero and $WDF$ provides a confidence of 1 meaning $100\%$. In all other cases $WDF$ provides a value between zero and 1.

\begin{figure}[t]
\centering
    \includegraphics[width=\linewidth]{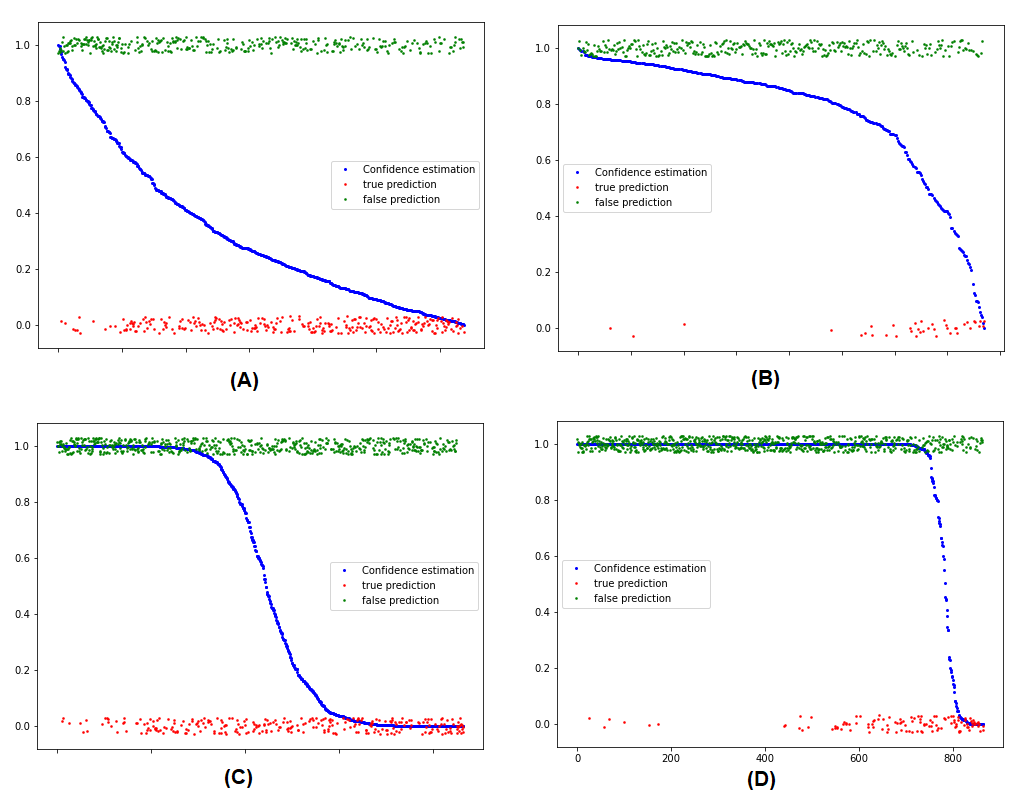}
    \caption{The figure demonstrates the behavior the confidence estimation, Equation  \ref{eq:wdf} (blue points). Each point corresponds to a single prediction and all the points are sorted according to their values. A red points signals a failing prediction whereas agreen point signals a successful prediction. In general less errors occur where the confidence estimation is high and vice versa. (A) and (B) correspond to the same prediction model for multilingual relation extraction. In (A) with German datasets the f-measure is 0.58 and (B) with Chinese the f-measure is 0.92. (C) and (D) correspond to two different image classification models applied on the same dataset. In (C), the model is trained on 5\% of the training data and has an f-measure of 0.68 and in (D), the model has been trained on 100\% of the training data and has an f-measure of 0.92.
    }
    \label{fig:confidence_estimation_general}
\end{figure}

Figure~\ref{fig:confidence_estimation_general} shows four examples of NN models having the behavior of the confidence estimators (in this case WDF, Equation \ref{eq:wdf}, but KRT shows also similar behavior). Each blue point corresponds to a particular classification (class prediction) and its value is the outcome of Equation \ref{eq:wdf}, when applied to the corresponding logits. The data points have been sorted along the x-axis according to their values. Each red point means that the prediction is failed. In analogy, each green point means a successful prediction. In all sub-figures, less errors occur where the confidence estimation is higher and vice versa, which means that the confidence estimator Equation (2) can predict the correctness of the model prediction. (A) and (B) correspond to the same prediction model for the semantic relation extraction, which is designed to operate in multilingual settings as well. If  (A) is applied on a German dataset the f1-score is 0.58 and if (B) applied on Chinese dataset the f1-score is 0.91. (C) and (D) correspond to two different image classification models applied on the same dataset. In (C), the first model (Image classification 5\%) was trained on 5\% of the training data to simulate a bad model and it achieved an f1-score of 0.68 and in (D), the model has been trained on 100\% of the training data and has an f1-score of 0.92. These fourexamples and many other example on different domains, technologies and datasets show similar behavior. That is the two estimators (Equations \ref{eq:krt} and \ref{eq:wdf}) behave indirect proportional to the probability of the prediction errors.

However, a function providing a value that is indirect proportional to the probability is not enough for the practical use, since we need to answerquestions in regards of  the meaning of a particular value of Equations  \ref{eq:krt} and \ref{eq:wdf} and how can we use it? Further questions consider whether the values universal, i.e. do the same values in two different models or two different datasets have the same meaning? How can one relate these values to a given confidence level? All these questions will be discussed in the following sections. The remainder of this paper is organized as follows: In the next section, we show how to use these functions (kurtosis and winner difference) in combination with thresholds. The thresholds are needed to identify those classifications that fulfill a particular confidence level.

\subsection{Thresholding the confidence values }
\label{sec:confidence_thresholds}
In this section, we show how to filter predictions to reach a given confidence level. To this end we discuss how to define and interpret thresholds in relation to the confidence estimators defined in Equations \ref{eq:krt} and \ref{eq:wdf}. From now, we will denote them simply thresholds. 

More specifically, we define a threshold $\mu$ to be a value that splits up the predictions obtained from a model into two subsets: The first subset, which we denote exploit \textit{exploit ($\mu$)}, contains predictions $x_i$ with $conf(x_i)>=\mu$ and the other subset, which we denote waste $\textit{waste ($\mu$)}$, contains the rest of the predictions $x_i$ having $conf(x_i)<\mu$.

\begin{figure}[t]
\centering
    \includegraphics[width=\linewidth]{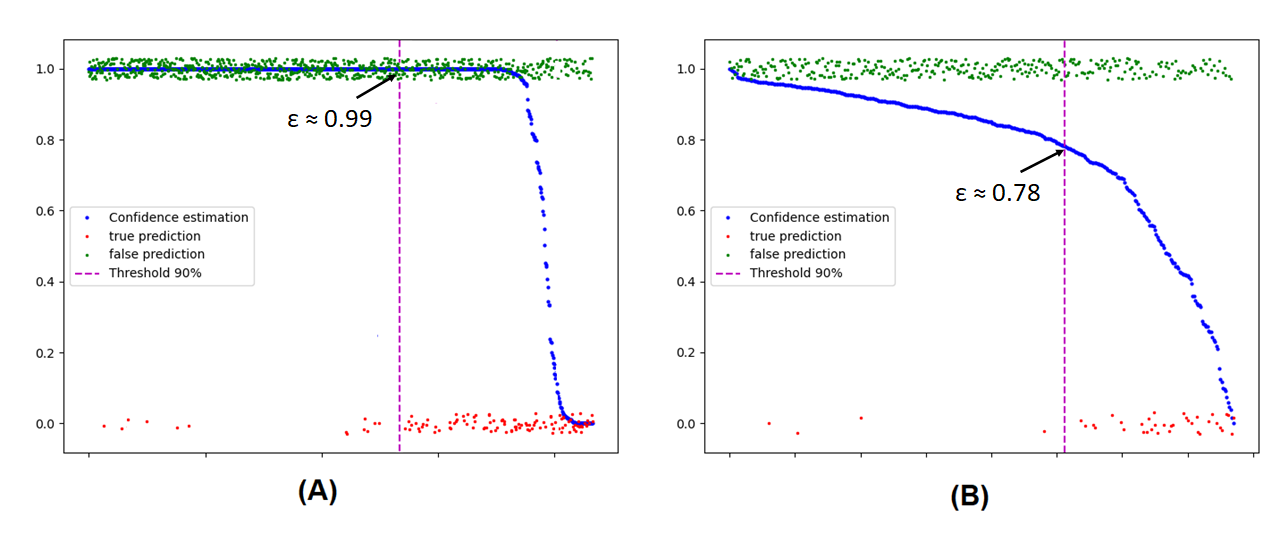}
    \caption{The figure shows how the confidence estimation values in different models correspond to different thresholds and vise versa. In (A) the threshold dividing the errors into $10\%$ and $90\%$ is $0.99$ while the same threshold in (B) it is $0.78$ (The two models are based on different technologies, but have the same accuracy - f-measure=$0.92$). (A) corresponds to a relation extraction model based on a pre-trained xlm-roberta model using Torch and (B) corresponds to an image classification model based on VGG16 pre-trained model using Keras.}
    \label{fig:thresholds}
\end{figure}

At first, let us check the universality of thresholds, in the sense whether the same threshold produces a similar error partitioning with different models. In other words, we want to know whether it is possible to define universal thresholds that produce repeatedly the same exploit ratio for  a given model. Figure~\ref{fig:thresholds} shows the confidence plots (based on Equation  \ref{eq:wdf} of two different models, (A) corresponds to a text relation extraction model based on a pre-trained xlm-roberta model using PyTorch and (B) corresponds to an image classification model based on VGG16 pre-trained model using Keras. The two models have the same accuracy (f-measure=$0.92$). Both thresholds plotted as vertical lines have been selected to accept only $10\%$ of the existing errors. We can see that the threshold are not similar in their values: $0.99$ in (A) and $0.78$ in (B).

The result in Figure~\ref{fig:thresholds}, which repeats itself in different models, also using Equation  \ref{eq:krt}, suggests that there a universal threshold can not be obtained. It means practically thresholds should be found each  model at hand. A topic which will be handled in the remainder of this section.

\begin{figure}[t]
\centering
    \includegraphics[width=\linewidth]{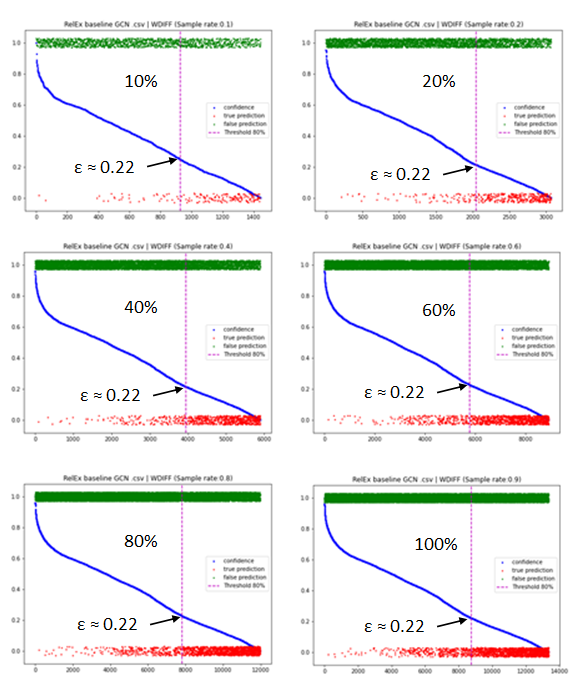}
    \caption{The figure shows empirically that for a given model, the shape of the confidence estimation tends to be constant for different samples of data. The same experiment has been performed on multiple datasets and repeated the same results.}
    \label{fig:sampling_thresholds}
\end{figure}

To find a threshold for a particular model, we use a test dataset, which exists always. The assumption is that the threshold is based on the model not on the data; To examine the role of the data  we defined an experiment to see whether a threshold depends on some data. The threshold behaves similar for other data as well as future data:: For a particular model, we created different subsets of the test set using random sampling (10\%, 20\%, ..., 100\% of the test set). For each sample, we calculated the threshold that accepts only 10\% of the existing errors. Figure~\ref{fig:sampling_thresholds} shows the results: Independent of the sample of the data, the calculated threshold remains unchanged, which means that the threshold is data-independent. The same results have been observed with different data and different models , more specifically with all the model and data combinations used in the Section~\ref{sec:evaluation}.

The result of this experiment is quite important and useful: Since the threshold is data independent with respect to error distribution, this means we can predefine a threshold for a given confidence level and a given model at training time based on the test data, which is normally available when creating a model. This important observation will be the basis of the next section, where we relate thresholds to confidence levels. Therefore, we want to mathematically formulate this observation as follows: The function $f(\mu)$ shows the relative error in exploit and is defined by 
\begin{equation}
   f(\mu) = \frac{e_{\mu}}{e_{\mu}+s_{\mu}}
   \label{eq:relative_errors_in_exploit}
\end{equation}
where $e_{\mu}$ and $s_{\mu}$ are the error predictions and successful predictions in $exploit(\mu)$ (i.e. instances $x$ satisfying $conf(x)>\mu$) is stable and data invariant for a given model, which has been empirically shown using various models and data. 

\subsection{Confidence Levels}
\label{sec:confidence_levels}
Until now, we considered the relative error in exploit (Equation~\ref{eq:relative_errors_in_exploit}) by finding a threshold that accepts a particular ratio of existing errors, e.g. $10\%$. Obviously, the final number of errors accepted by filtering with threshold$\mu$ depends on the original number of errors, which consequently depends on the accuracy of the model. In this section, we introduce a method for providing a threshold that can be used as a filter to generate a subset of predictions that fulfills a given accuracy regardless of the original accuracy. In other words, given a confidence level and a model with an accuracy, we need to define a threshold that ensures an accuracy sufficient for the confidence level, i.e. a threshold that defines the optimal exploit/waste ratio, in such a way that the wished accuracy is however met.

Let p be the original accuracy of the model $M$ based on evaluation using a test set T. We want to find a threshold $\mu$ that produces an exploit with accuracy $q$ where $q>p$ (Note that when q<p, no filtering is required, because accuracy is already optimized). For the sake of simplicity, we assume that the accuracy is measured by by the number of correct predictions divided by the total number of predictions. Let $e_{\mu}$ and $s_{\mu}$ be the number of wrong predictions and true predictions in $exploit(\mu)$ respectively, then

\begin{equation}
   q = \frac{s_{\mu}}{e_{\mu}+s_{\mu}} = 1 - f(\mu)
   \label{eq:threshold_for_a_confidence_level}
\end{equation}
which is characteristic for a given model according to Equation \ref{eq:relative_errors_in_exploit}. Now, the task is to find the threshold $\mu$ that meets Equation \ref{eq:threshold_for_a_confidence_level}. Since the percentage of the errors before filtering is $1-p$ and the maximum error percentage we aim to reach is $1-q$, we propose the threshold satisfying Equation \ref{eq:threshold_for_a_confidence_level} to be $\mu$, where 
\begin{equation}
  e_{\mu} =  e  \frac{1-q }{1-p} 
   \label{eq:final_threshold}
\end{equation}
where $c$ is the total number of error and $c_{\mu}$ is the accepted number of error, which means $\mu$ should be experimentally selected using the testset, such that the exploit contains $ e_{\mu}$ errors and the waste contains $c -  e_{\mu}$ errors.

\section{Evaluation}
\label{sec:evaluation}
\begin{figure*}[ht]
\centering
    \includegraphics[width=\linewidth]{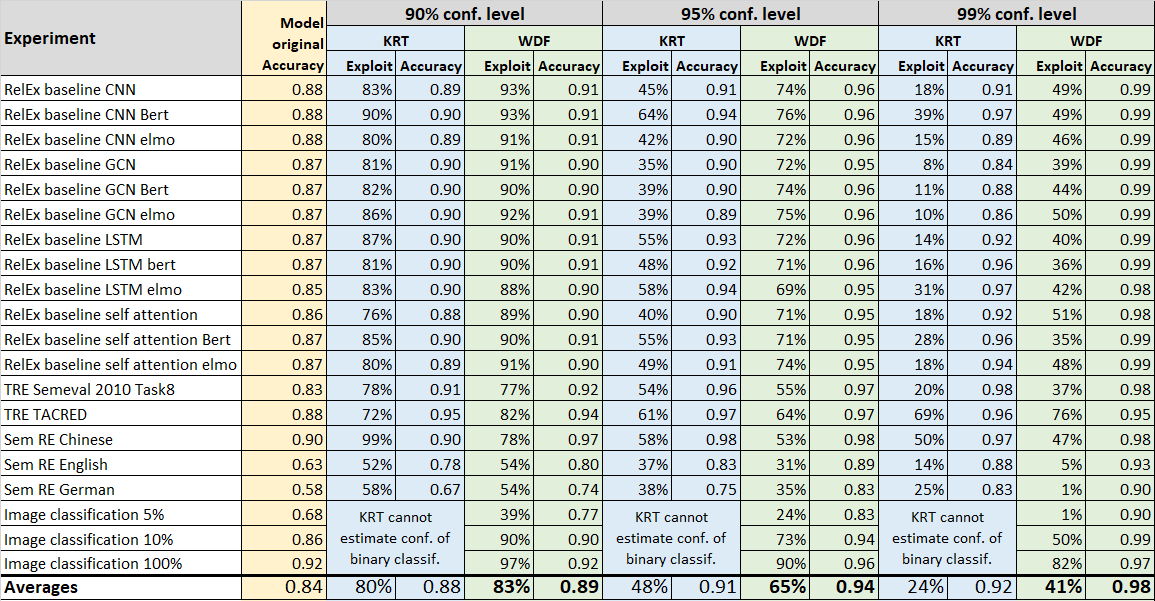}
    \caption{Evaluation results of the confidence estimation functions KRT (Equation~\ref{eq:krt}) and WDF (Equation~\ref{eq:wdf}) on 20 different models in three difference confidence levels, $90\%$, $95\%$ and $99\%$. The evaluation considers for each confidence level the accuracy in f-measure on the exploit compared with the original accuracy (the second column) and the exploit ratio in percent. The last three experiments are binary (two classes) for which KRT doesn't work.}
    \label{fig:results}
\end{figure*}
The two confidence estimation functions proposed in Section~\ref{sec:confidence_functions}, namely $KRT$ (Eq.~\ref{eq:krt}) and $WDF$ (Eq.~\ref{eq:wdf}) have been intensively tested using various models of different technologies in different domains and different datasets. 

No evaluation against state-of-the-art methods has been performed, simply because to our knowledge no methods exists in the state-of-the-art that estimates the confidence of a particular prediction.

Rather, we evaluated the methods by comparing the original accuracy of the model with its accuracy on the exploited relations, i.e. after performing filtering by using the confidence threshold. This has been done for different confidence levels as well. Also, we have considered the exploit ratio for the evaluation. In particular, we used the following measures:

\begin{itemize}
\item \textbf{Enhanced accuracy (E.ACCU):} The accuracy of the models measured on the exploit, compared with the original accuracy (ACCU) on the dataset. That is the accuracy when excluding predictions with confidence value lower than a threshold $\mu$, calculated according to Equation~\ref{eq:final_threshold} for the given confidence level.
\item \textbf{Exploit ratio (EXPL.R)}: That is the ratio of predictions having confidence values over the threshold $\mu$, i.e. the ratio of the predictions recommended to be used for the given confidence level.    
\end{itemize}

We used 20 models of different technologies performed on different datasets as well as different domains, such as imaging, relation extraction and semantic domain. The aim was to ensure that the methods work for a broad variety of models, technologies and data. In particular, we used the following groups of experiments:
\begin{enumerate}
    \item \textbf{RelEx / Relation extraction on TACRED dataset}: In this category, we included 12 models from the RelEX benchmark \cite{RelEx}, which includes NN-based relation extraction algorithms based on various combinations of state-of-the-art embedding systems like BERT and ELMO as well as different NN technologies like CNN (convolutional neural network), GCN (graph convolutional network) and self attention. All experiments in this group were performed on the TACRED dataset \cite{TACRED_dataset}, where predefined relations between named entities are to be identified.
    \item \textbf{TRE / Relation extraction on TACRED dataset}: The TRE \cite{DFKI_1} is an NLP-based algorithm that uses pre-trained language representations to improving Relation Extraction. It was used in combination of the TACRED dataset \cite{TACRED_dataset}.
    \item \textbf{TRE / Relation extraction on SemEval2010 Task 8}: This is the sasme algorithm described in 2), but applied on the SemEval1010 Task 8 \cite{SemEval_dataset}.
    \item \textbf{TransRelation / Semantic relation on CogALex}: The CogALex VI Task \footnote{https://sites.google.com/site/cogalexvisharedtask/} is a shared task for semantic relation extraction, i.e. to discover whether semantic relations between words in a text exist and which. The TransRelation \cite{Transrelation} algorithm uses the pre-trained model XLM-RoBERTa to build a semantic relation extraction model. Note that this kind of relation extraction differ from the previous groups: while the previous groups have to do with specific relations like producer (e.g. Siemens produces Motors), this groups deals with general semantic relation like synonym and Antonym (e.g. the words wide and far are synonyms).
    \item \textbf{Named entity recognition}: xxxxxxx xxxxx xxxxxxx xxxxxxxxxxxxxx xxxxxxxxx xxxxx xx xxx xxx xx xxx xxxxxx xxxx xxxxxxxx xxxxxxxx xxx xxxxxxx xxxxx
    
    \item \textbf{VGG16 / Image classification}: This experiment group includes binary image classification models based on the pre-trained VGG16 model \cite{vgg16}, which are performed on the known dogs and cats image dataset~\cite{asirra}. The first goal of including this experiment group is to show that the methods are applicable on other domains and the performance of the confidence estimation repeats itself on domains other than text. The second goal is to show that the WDF confidence estimator works with any number of classes, even binary classification.
\end{enumerate}

Figure~\ref{fig:results} shows the results of the evaluation. The first columns show the name and original accuracy of the model. For each of the confidence levels $90\%$, $95\%$ and $99\%$, it shows the enhanced accuracy (E.ACCU) and the exploit ratio (EXPL.R) for both confidence estimation functions KRT and WDF. The evaluation results can be summarized as follows: 
\\
     \textbf{Significant increase of accuracy:} The E.ACCU is significantly higher than the original accuracy. Note that if the original accuracy is high compared with the confidence level (e.g Experiments xxx, yyy, zzz), the exploit ratio is very high and the accuracy is almost unchanged. This has been expected and actually already planned in the threshold calculation (Equations~\ref{eq:threshold_for_a_confidence_level} and \ref{eq:final_threshold}) since the threshold is set based on the difference between the original accuracy in F1-measure and the confidence level. Higher increase of accuracy is observed with the less accurate models (e.g. Experiments numbers xxx,yyy,zzz). As designed, the threshold tends to keep the accuracy close to the given confidence level regardless of the original accuracy.
\\
     \textbf{Exploit:} The EXPL.R values looks reasonable except in extreme cases, i.e. when the model accuracy is very low and the confidence level is very high, e.g. Experiments xxx,yyy,zzzz. In this case the exploit drops very much.
\\    
    \textbf{WDF outperforms KRT:} In most experiments in all confidence levels, WDF outperforms KRT both in accuracy and exploit ratio. This provides an evidence to make WDF (Equation~\ref{eq:wdf}) the recommended confidence estimation function.

\section{Conclusion}
This paper provides methods to infer and increase the confidence in classifications generated by a neural network (NN). The methods don't require any changed on existing NN. In particular, the contribution of this paper are: 
\begin{itemize}
    \item Two confidence estimation functions that, given a prediction, they provide estimates for the correctness of the prediction, i.e. estimate of the likelihood that the prediction is correctly classified. This estimate is inferred based on the distribution of the logits corresponding to the underlying prediction.
     \item A method that, given a classification model and a confidence level, calculates a threshold that separates the classifications generated by this model into two subsets, where one of them fulfills the accuracy required by the given confidence level to be used.
\end{itemize}
The provided methods are recommended for filtering predictions generated in the process of knowledge extraction, like named entity recognition and relation extraction, e.g. based on web Scraping. In those applications there is availability in data, but lack in model accuracy and generalization due data noise and heterogeneity. Our methods help in solving the problem by increasing the confidence by filtering on cost of recall which is less important due the data availability. 

\bibliography{references}
\bibliographystyle{plain}

\end{document}